%% file: paper.tex
\documentclass[]{jingdong}
\usepackage[toc,page,header]{appendix}
\input{common}

\usepackage{cleveref}
\theoremstyle{plain}

\theoremstyle{definition}

\theoremstyle{remark}
\usepackage{subcaption}

\usepackage[textsize=tiny]{todonotes}
\usepackage{fontawesome}

\title{JoyAgent-JDGenie: Technical Report on the GAIA}

\affiliation{JINGDONG CHO-EI Team}

\input{abstract}

\date{\today}
\checkdata[Code]{\url{https://github.com/jd-opensource/joyagent-jdgenie}}

\begin{document}
\maketitle

\input{intro}
\input{related}
\input{method}
\input{eval}

\input{conclusion}
\input{acknowledge}
\clearpage
\bibliographystyle{plainnat}
\bibliography{cite}

\input{appendix}
\end{document}

%% file: common.tex
\usepackage{latexsym}
\usepackage[T1]{fontenc}
\usepackage[utf8]{inputenc}
\usepackage{microtype}
\usepackage{inconsolata}
\usepackage{graphicx}
\usepackage{hyperref}       
\usepackage{url}            
\usepackage{booktabs}       
\usepackage{amsfonts}       
\usepackage{nicefrac}       
\usepackage{stackengine}
\usepackage{microtype}      
\usepackage{colortbl}
\usepackage{xcolor}
\usepackage{amsmath}
\usepackage{amssymb}
\usepackage{amsthm}
\usepackage{mathrsfs}
\usepackage{pifont}
\usepackage{MnSymbol}
\usepackage{balance}
\usepackage{enumitem}
\usepackage{listings}
\usepackage{xcolor}
\usepackage{natbib}
\usepackage{multicol}

\AtBeginDocument{%
  \providecommand\BibTeX{{%
    \normalfont B\kern-0.5em{\scshape i\kern-0.25em b}\kern-0.8em\TeX}}}

\makeatletter
\DeclareRobustCommand\onedot{\futurelet\@let@token\@onedot}
\def\@onedot{\ifx\@let@token.\else.\null\fi}

\usepackage{setspace}
\usepackage{mathtools}

\usepackage{multirow,booktabs}
\usepackage{subcaption}

\newcommand{\gaia}{\textit{GAIA}}

\newcommand{\owo}[1]{\textsc{OAgents}}

\definecolor{lightgreen}{RGB}{144, 238, 144} 
\definecolor{lightred}{RGB}{255, 105, 97}

\newcommand{\fakeparagraph}[1]{\vspace{1mm}\noindent\textbf{#1.}}

\newtcolorbox{promptbox}[2][Prompt]{
colback=black!5!white,
arc=5pt, 
boxrule=0.5pt,
fonttitle=\bfseries,
title=#1, 
before upper={\small}, fontupper=\fontfamily{ptm}\selectfont,
colframe=#2, 
}
\definecolor{ogreen}{RGB}{34, 139, 34}

%% file: abstract.tex
\abstract{

Large Language Models (LLMs) are increasingly deployed as autonomous agents for solving complex real-world tasks. Yet, existing systems often emphasize isolated improvements—such as expanding toolkits, refining prompts, or adjusting planning heuristics—without a unifying framework to ensure robustness, adaptability, and reproducibility. In this work, we present a generalist agent architecture that integrates three core components: (i) a collective multi-agent framework of \texttt{Plan–Execute} and \texttt{ReAct} agents coordinated through critic model voting, balancing stability with flexibility; (ii) a hierarchical memory system combining working, semantic, and procedural layers to enable long-horizon continuity and adaptive control; and (iii) a refined tool suite focused on search, code execution, and multimodal parsing, exposed through schema-consistent and auditable interfaces. Evaluated on the GAIA benchmark~\cite{mialon2023gaia}, our framework achieves 75.2 Pass@1 and 82.4 Pass@3 on validation, and 67.1\footnote{\url{https://huggingface.co/spaces/gaia-benchmark/leaderboard}} Pass@1 on test set, surpassing open-source baselines and approaching the performance of proprietary systems. These results underscore the importance of system-level integration for advancing generalist agents. Beyond immediate gains, our work highlights a path toward scalable, resilient, and adaptive AI assistants capable of operating across domains and tasks.}

%% file: intro.tex
\section{Introduction}

The pursuit of Artificial General Intelligence (AGI) increasingly centers on agents: autonomous systems that can plan, reason, and act across diverse real-world tasks. Unlike conventional large language models (LLMs), which excel primarily at text generation, agentic systems must integrate reasoning with external actions, leverage memory across interactions, and adapt dynamically to unforeseen contexts. This shift positions LLM-based agents not just as conversational tools, but as general-purpose assistants capable of problem solving in open-ended, multi-modal, and continuously evolving environments.

Over the past two years, a wide range of agent frameworks have emerged. Early role-based multi-agent systems such as CAMEL~\cite{li2023camel} and MetaGPT~\cite{hong2024metagpt} demonstrated that structured collaboration could elicit disciplined reasoning behaviors, while more recent initiatives like OAgents~\cite{oagent}, AWorld~\cite{aworld2025}, Alita~\cite{qiu2025alita}, and OWL~\cite{owl2025} highlight new design philosophies: empirical scaling studies, co-evolutionary training loops, meta-tool creation, and modular planners. Benchmarks such as \gaia{}~\cite{mialon2023gaia} have revealed both the promise and the limitations of these systems—while they achieve notable progress, many approaches remain brittle, either due to over-reliance on hand-crafted workflows or lack of stability under real-world uncertainty.

\label{sec:intro}

\begin{figure}[t]
    \centering
    \includegraphics[width=0.8\linewidth]{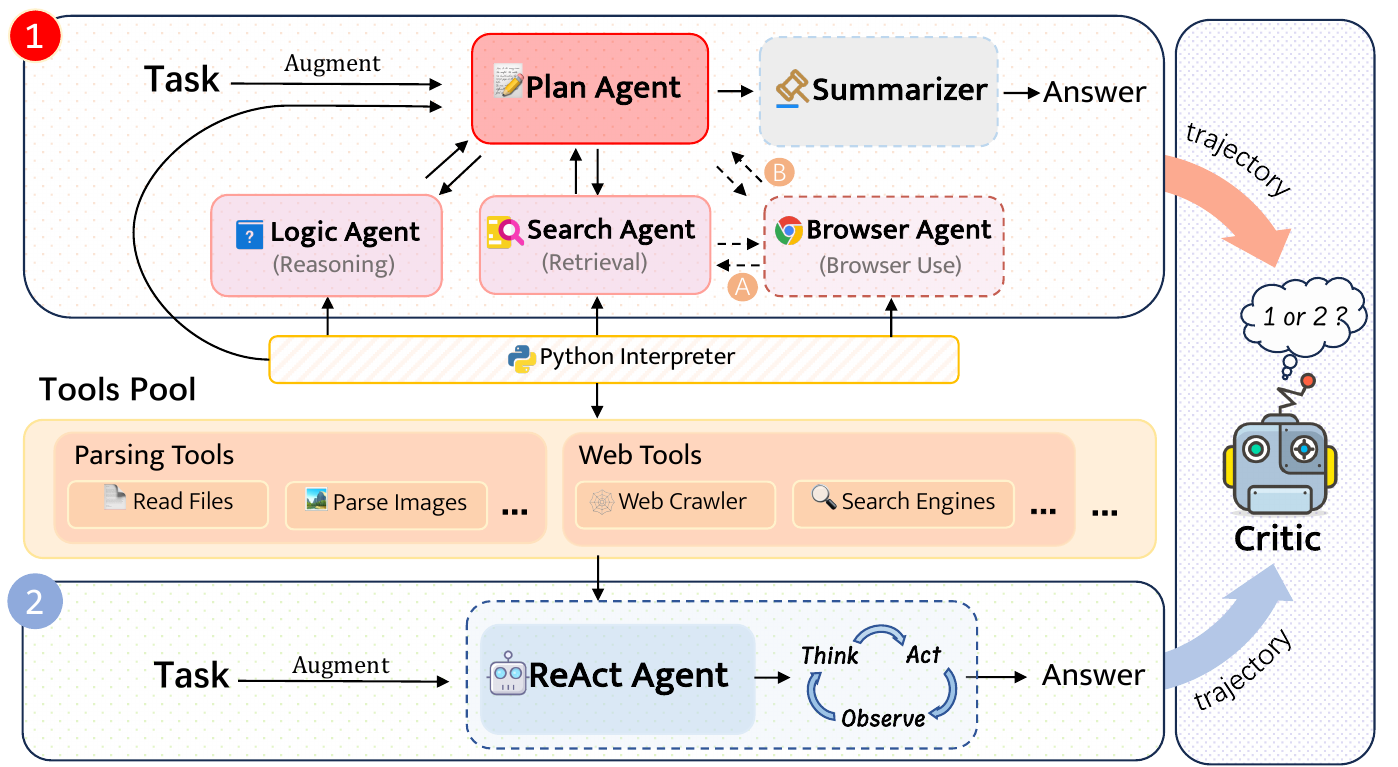}
    \caption{The overview of the fusion agent architecture.}
    \label{fig:overview}
\end{figure}

In this work, we address these challenges by proposing a system-level framework that integrates three core components: (i) a heterogeneous ensemble of agents spanning both \texttt{Plan–Execute} and \texttt{ReAct} paradigms, coordinated through posterior voting to balance reliability and adaptability; (ii) a hierarchical memory design that combines working, semantic, and procedural layers to enable long-horizon continuity and adaptive control; and (iii) a refined tool ecosystem emphasizing search, code execution, and multimodal parsing, each wrapped into schema-consistent and auditable interfaces. Together, these design choices yield robust performance gains on the GAIA benchmark, setting state-of-the-art results among open-source systems and narrowing the gap to proprietary frameworks.

%% file: related.tex
\section{Related Work}
\label{sec:related}

The study of LLM-based agents has progressed from role-specialized collaborations to increasingly modular systems oriented toward general assistance. Early frameworks such as CAMEL~\cite{li2023camel} and MetaGPT~\cite{hong2024metagpt} showed that assigning distinct roles and procedures elicits more disciplined reasoning, especially in structured domains like software engineering. As evaluations moved toward open-ended tasks, recent efforts emphasized broader empirical analysis and unified infrastructures: OAgents~\cite{oagent} systematically studies design choices for effective agents at scale, while AWorld~\cite{aworld2025} provides a unified playground that supports both computer- and phone-use tasks, encouraging iteration on the full stack of planning, tools, and evaluation. In parallel, Alita~\cite{qiu2025alita} advances the idea that tools themselves can be dynamically constructed and composed, reducing reliance on heavy predefinition and enabling self-evolution of agent capabilities. Complementary platforms (e.g., AutoAgent~\cite{tang2025autoagent}) lower the barrier to assembling agents and workflows, making orchestration more accessible without sacrificing modularity.

A second thread concerns the substrate that turns models into reliable systems: tools, retrieval, and memory. Toolformer~\cite{schick2023toolformer} demonstrated that models can self-learn to call APIs, and surveys such as Qin et al.\cite{qin2024tool} underscored that robust tool interfaces and execution traces are core to stability. Hierarchical research agents (e.g., Open Deep Research\cite{opendeepresearch} and DeepResearchAgent~\cite{DeepResearchAgent}) pair retrieval-centric planning with iterative decomposition, while memory mechanisms progress from reflective feedback~\cite{shinn2023reflexionlanguageagentsverbal} to hierarchical designs like A-Mem~\cite{xu2025mem} that separate working, semantic, and procedural layers—supporting long-horizon continuity and adaptive control. These system utilities collectively motivate architectures where communication is structured, tool calls are auditable, and prior experience can be retrieved and reused.

Finally, generalization across domains has become a central goal. OWL~\cite{owl2025} tackles this by decoupling domain-agnostic planning from domain-specific execution (WORKFORCE), training only the planner—via supervised trajectories and reinforcement learning—to transfer across new environments without retraining worker agents. Orthogonal efforts (e.g., AgentRefine~\cite{fu2025agentrefine}, TapeAgents~\cite{bahdanau2024tapeagentsholisticframeworkagent}, and MiroFlow~\cite{miroflow}) study reproducibility, refinement, and stability at scale. Benchmarks such as GAIA~\cite{mialon2023gaia} and BrowseComp~\cite{wei2025browsecomp} crystallize these trends by testing multimodal reasoning, browsing, and execution, revealing that while proprietary frameworks (e.g., Deep Research~\cite{deepresearch}, h2oGPTe~\cite{h2oGPTe2024h2oGPTe}) are strong, open-source systems are rapidly closing the gap. Distinct from the above, our work contributes a system-level design that intentionally fuses complementary agentic paradigms with hierarchical memory and a statistically validated tool suite, aiming for robust gains under real-world constraints.

%% file: method.tex
\section{Architecture}
\label{sec:method}
Building effective generalist agents requires more than scaling language models: it demands a careful integration of planning strategies, memory mechanisms, and tool infrastructures into a coherent system. Prior work has shown that isolated improvements in any one component—such as adding new tools, refining prompts, or adjusting planning heuristics—often lead to limited or unstable gains. Our approach instead emphasizes system-level design, where diverse agent paradigms, structured memory hierarchies, and statistically validated tool sets are woven into a unified framework. This section details our methodology, beginning with agent architectures, followed by memory design, and concluding with tool integration.

\subsection{Agents}
We design our framework around a heterogeneous ensemble of agents that reflects two complementary paradigms of agentic reasoning. The first follows the \texttt{Plan–Execute} principle, where a high-level plan is generated in advance, executed step by step, and periodically revised through lightweight reflection. This structure provides a low-variance pipeline, ensuring that tasks with deterministic decomposition can be executed reliably. The second follows the \texttt{ReAct} paradigm, in which reasoning and action are tightly interleaved, allowing the agent to replan dynamically at every step. Although this style exhibits higher variance, it excels in exploratory tasks that require adaptive reasoning under uncertainty.

To reconcile these statistical trade-offs, we implement a hierarchical multi-agent ensemble. A Supervisor Agent built upon the \texttt{Plan–Execute} framework ensures global coherence of the solution trajectory, while multiple Single Agents based on \texttt{ReAct} provide step-level adaptability. At inference, their outputs are aggregated through posterior voting, which can be configured with 3 or 5 models depending on resource availability. For instance, a three-way ensemble may combine two \texttt{ReAct} agents with a \texttt{Plan–Execute} agent. Empirically, this mixture consistently improves pass rates across GAIA tasks, highlighting the benefit of balancing bias and variance through ensemble decision-making.

Inter-agent interaction is governed by a structured communication protocol. Each agent produces not only a candidate solution but also a message object that records reasoning chains, tool invocations, and intermediate evidence. These messages are transmitted through a central communication hub, stored in the working memory buffer, and made accessible to other agents for cross-validation. By constraining communication to structured formats rather than free-form dialogue, we ensure consistency and prevent uncontrolled conversational drift. This architecture resembles a cooperative yet disciplined debate, where agents can critique or support one another proposals, leading to more reliable outcomes.

\subsection{Memory}

Memory is a central component that underpins both long-term continuity and short-term adaptability of our framework. We design a hierarchical memory system consisting of three layers.

\begin{itemize}
    \item \textbf{Working Memory}
    stores the live execution context, including current plans, intermediate tool outputs, and exchanged messages between agents.
    \item \textbf{Semantic Memory}
    records the trajectory of completed tasks, including successes, failures, and decision rationales, compresses episodic traces into distilled knowledge units via summarization and embedding, ensuring that relevant lessons can be retrieved even when raw histories are prohibitively long.
    \item \textbf{Procedural Memory}
     is embedded in the form of finely tuned system prompts. These prompts encode guidelines such as how to prioritize information sources, when to replan, or how to handle conflicting evidence. Unlike static instruction prompts, our procedural memory is dynamically adjusted based on accumulated experience.
\end{itemize}

During inference, retrieval from long-term memory is mediated by semantic similarity search, ensuring that agents can access precedent cases that align with the current task. Retrieved items are then injected into working memory as auxiliary context. In addition, procedural memory functions as a meta-controller, shaping how agents interpret retrieved traces and adapt their planning strategies. This combination enables unbounded historical continuity, where the agent retains identity and knowledge across arbitrarily extended interactions without overwhelming the underlying LLM backbone.

\subsection{Tools}
Tool design and integration form the backbone of the agent’s factual acquisition capacity. Rather than maximizing the number of available tools, we identify and refine the classes of tools that statistically contribute most to task success. Our final tool suite centers around three categories: search, code execution, and local multimodal parsing.

The search subsystem employs a multi-source aggregator that queries \texttt{Google}, \texttt{Bing} and \texttt{DuckDuckGo}, supplemented with domain-specific interfaces such as \texttt{Wikipedia} search, \texttt{Arxiv} advanced retrieval, and multiple \texttt{GitHub} search APIs. To avoid brittle dependence on a single provider, queries are reformulated via a reflection–expansion loop, where the agent first analyzes ambiguities, then generates alternative formulations with morphological and semantic variants. Retrieved documents are parsed using a minimalist browsing tool set restricted to Search, Visit, and Read, which reduces error propagation from overly complex navigation.

The code execution environment is implemented as a secure \texttt{Python} sandbox. Tool calls follow a uniform API: the agent generates structured code snippets, which are executed in isolation, and the execution traces are automatically stored in working memory. This allows downstream agents to reason not only over outputs but also over execution logs, supporting trace-based debugging and iterative refinement.

For multimodal parsing, we introduce 17 specialized interpreters capable of handling PDFs, spreadsheets, presentations, audio, video, and image files. Each interpreter exposes a lightweight interface (e.g., \texttt{parse\_pdf}, \texttt{extract\_audio\_transcript}, \texttt{analyze\_image}) that returns structured outputs rather than free-form text, enabling downstream reasoning to operate on consistent schemas. Crucially, these interpreters integrate directly into the communication hub: parsed content is added to working memory, making it accessible to all collaborating agents.

The combination of carefully chosen search, code, and multimodal tools results in 30–60\% gains on \texttt{Plan–Execute} baselines, establishing a robust substrate upon which more advanced ensemble and reinforcement strategies can be layered.

%% file: eval.tex
\section{Experiments} \label{sec:eval}
\subsection{Experimental Setting}
\fakeparagraph{Dataset}
\gaia{}\cite{mialon2023gaia} is a benchmark designed to evaluate general-purpose AI assistants through 300 test and 165 validation real-world, scenario-based questions covering daily tasks, tool usage, reasoning, multimodal inputs, and web browsing. While these tasks may appear straightforward for humans, they remain highly challenging for advanced AI systems. Each question is associated with a unique ground-truth answer, and model performance is measured using exact match accuracy.

\fakeparagraph{Metrics} 
We adopt the evaluation protocol of the \gaia{} benchmark, which relies on exact match accuracy. The main metric is \textit{Pass@N}, defined as the probability that at least one correct solution appears among $N$ independent execution. This metric, commonly used in tasks like code generation, captures whether the model can generate a valid solution at least once. Unless otherwise specified, our experiments report the average \textit{Pass@1} score, indicating the model’s ability to produce a correct answer in a single task run.

\fakeparagraph{Baselines}
For a comprehensive evaluation, we compare our system against various baselines from three primary types: agentic models (Search-o1-32B, WebThinker-32B, WebDancer-32B, WebShaper-32B); closed-source frameworks (Langfun~\cite{Peng_Langfun_2023}, TraseAgent~\cite{trase2024trase}, Deep Research~\cite{deepresearch}, h2oGPTe~\cite{h2oGPTe2024h2oGPTe}, and Desearch~\cite{desearch}); and open-source systems (OWL~\cite{owl2025}, TapeAgent~\cite{bahdanau2024tapeagentsholisticframeworkagent}, AutoAgent~\cite{tang2025autoagent}, Open Deep Research~\cite{opendeepresearch}, Smolagents~\cite{smolagents}, OAgent~\cite{oagent}, and MiroFlow~\cite{miroflow}). This selection captures a broad range of the latest developments in both proprietary and open multi-agent systems, establishing a robust benchmark for assessing the performance.

\subsection{Main Results}
The results presented in Table 1 offer several key insights into the performance of various agent frameworks on the \gaia{} benchmark. Our proposed method achieved an average score of 75.2 at Pass@1 and 82.4 at Pass@3, demonstrating competitive results against all other evaluated frameworks, including both closed-source and open-source alternatives. This outcome underscores the robustness and effectiveness of our agent's design.

For Level 1 tasks, our method achieved a score of 86.8, matching the top-tier performance and demonstrating the reliability of our low-level agents and their underlying system utilities. Compared to leading closed-source solutions such as Langfun (71.5) and MiroFlow (74.5), our approach shows substantial improvements in overall average performance and maintains superior accuracy across both Level 1 and Level 2 tasks. Notably, the highest-performing solutions predominantly leverage Claude-family models, underscoring the importance of foundation model selection. These results collectively validate the effectiveness of our framework for general-purpose agent applications.

We publicly run our agent against GAIA testset, and obtain a relatively high score of 67.1. Please refer to GAIA's official leaderboard\footnote{\url{https://huggingface.co/spaces/gaia-benchmark/leaderboard}}. Regarding Open Deep Research ~\cite{opendeepresearch} and Smolagents \cite{smolagents}, their reported results were directly adopted from OAgents \cite{oagent} due to the substantial computational resources required for replication.

\begin{table}[!thb]
\centering
\caption{Performance of various agent projects on the GAIA benchmark.}
\label{tab:related}
\begin{tabular}{lcccccl} 
\toprule
\midrule
\textbf{Framework} & \textbf{Pass@1} & \textbf{Pass@3} & \textbf{Level 1} & \textbf{Level 2} & \textbf{Level 3} & \textbf{Model Family} \\
\midrule
\multicolumn{7}{l}{\cellcolor[RGB]{235, 228, 230}{\emph{Agentic Model}}} \\
\midrule
Search-o1-32B
& 39.8 & & 53.8 & 34.6 & 16.7 & QwQ-32B \\
WebThinker-32B
& 48.5 & & 56.4 & 50.0 & 16.7 & QwQ-32B \\
WebDancer-32B
& 51.5 & 64.1 & 61.5 & 50.0 & 25.0 & QwQ-32B \\
WebShaper-32B
& 53.3 & 61.2 & 69.2 & 50.0 & 16.6 & QwQ-32B \\
\midrule
\multicolumn{7}{l}{\cellcolor[RGB]{235, 216, 221}{\emph{Closed-source Agent Frameworks}}} \\
\midrule
Langfun
& 71.52 & & 83.02 & 68.60 & 57.69 & Claude-3-7 etc.\@ \\
TraseAgent
& 70.30 & & 83.02 & 69.77 & 46.15 & Claude etc.\@ \\
DeepResearch
& 67.36 & & 74.29 & 69.06 & 47.60 & - \\
h2oGPTe
& 63.64 & & 67.92 & 67.44 & 42.31 & Claude-3.5  \\
Desearch
& 56.97 & & 71.70 & 58.14 & 23.08 & GPT-4o \\
\midrule
\multicolumn{7}{l}{\cellcolor[RGB]{235, 181, 195}{\emph{Open‐source Agent Frameworks}} }\\
\midrule
OWL
& 69.1 &  & 84.9  & 67.4 & 42.3 & Claude-3-7 etc.\@ \\
TapeAgents
& 55.8 & & 71.7 & 53.5 & 30.8 & Claude-3-7 etc.\@ \\
AutoAgent
& 55.2 & & 71.7 & 53.4 & 26.9 & Claude-3-5 etc.\@ \\
Open Deep Research
& 55.2 & & 67.9 & 53.5 & 34.6 & OpenAI o1 \\
Smolagents
& 49.7 & & 54.7 & 53.5 & 26.9 & Openai o1 etc.\@ \\
OAgent
& 66.7 & 73.9 & 83.0 & 74.4 & 53.9 & Claude-3-7 etc.\@ \\
MiroFlow
& 74.5 & 82.4 & - & - & - & Claude-3-7 etc.\@ \\
\midrule
\textbf{Ours}
& \textbf{75.2} & \textbf{82.4} & 86.8 & 77.9 & 42.3 & Claude-4 + o4-mini\@ \\
\midrule
\bottomrule
\end{tabular}
\end{table}

\subsubsection{Exploratory Evaluations}
\noindent \textbf{Agent Pattern}

We experimented with various Agent patterns, including Multi-Agent with Plan-Executor and Single Agent with ReAct~\cite{yao2023react}, corresponding to Circle 1 and Circle 2 in Fig~\ref{fig:overview}, respectively.

For the Single Agent approach, we adopted a basic ReAct pattern, providing all tools (excluding browser-use tools) and increasing the maximum execution steps. Surprisingly, this simple structure did not exhibit performance collapse; instead, it achieved the highest performance of 71.5 under the non-fusion approach. While its performance on Level 3 problems was lower than other MultiAgent methods, its superior performance on Level 1 problems improved the overall average performance.

\begin{table}[!h]
\centering
\setstretch{1.1}
\caption{Performance comparison of different system architecture on \gaia{} benchmark. The Fusion refers to fusing Single and Multiple (3) with an additional critic model.}
\label{tab:pattern}
\vspace{-9pt}
\begin{tabular}{l|cccc}
\toprule
Model & Average & Level 1 & Level 2 & Level 3  \\
\midrule
Single       & 71.5 & 84.9 & 75.6 & 30.8 \\
Multiple (2) & 69.9 & 80.6 & 74.4 & 33.5 \\
Multiple (3) & 70.3 & 81.1 & 74.4 & 34.6 \\
Multiple (4-A)& 52.7 & 56.6 & 58.1 & 26.9 \\
Multiple (4-B)& 58.8 & 60.3 & 65.1 & 34.3 \\
Fusion & 75.2 & 86.6 & 77.9 & 42.3 \\
\bottomrule
\end{tabular}
\end{table}

For the Multi-Agent approach, we constructed four different types of agents with distinct roles: a Plan Agent responsible for high-level task planning, a Retrieval Agent for web information retrieval, a Logic Agent for complex reasoning and code generation, and a Browser Agent for web page interaction. Different agents register different tools according to their roles. Additionally, all agents operate as CodeAgent, completing tool usage and inter-agent communication through Python code execution. By combining different agents, we formed Multi-Agent systems with various architectures. Specifically, Multiple (2) represents Plan + Retrieval Agent, Multiple (3) represents Plan + Retrieval + Logic Agent, and Multiple (4/5) represents systems using all agents with structures shown in Fig~\ref{fig:overview}. We found that without browser-use, MultiAgent can significantly improve performance on Level 3 problems, but performance degrades on simple Level problems. After introducing the browser agent, the system performance exhibits significant deterioration.

For the Fusion method, we selected Single Agent and Multiple (3) to combine the advantages of both architectures. We additionally incorporated a Critic model that performs comparative analysis of execution trajectory segments from both systems and provides final answers while strictly adhering to answer formats. As shown in Table~\ref{tab:pattern}, the fusion approach achieved more accurate generation through comparative analysis.

\noindent \textbf{Base Model}

LLMs serve as the brain of Agent systems, and we compared the impact of different foundation models on Agent system performance. Since we adopted the CodeAgent execution approach, the results reflect coding capabilities rather than agentic tool-calling abilities. Consistent with results from other open-source frameworks, we found that models from the Claude family performed best, with Claude-4-sonnet and Claude-3.7-sonnet achieving average scores of 75.2 and 68.3, respectively. Additionally, the thinking model o4-mini outperformed the non-thinking model gpt-4.1.

\begin{figure}[t]
    \centering
    \includegraphics[width=0.5\linewidth]{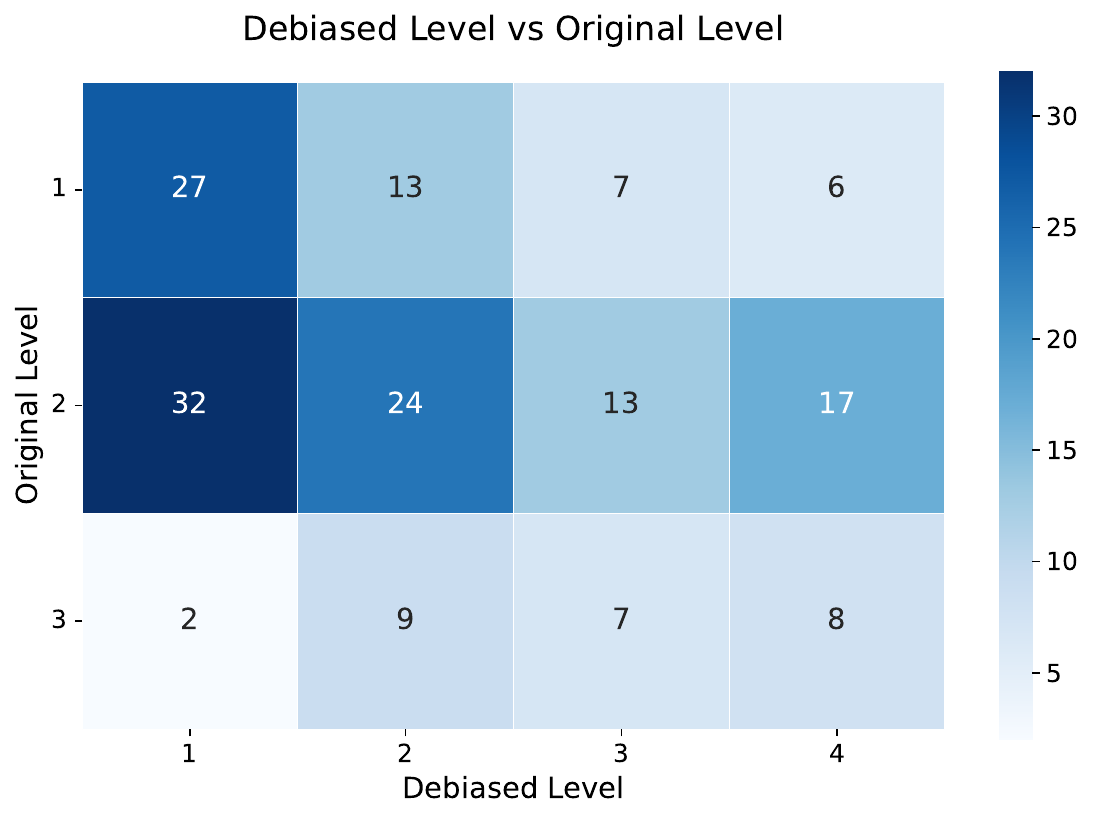}
    \caption{The distribution between original level and reassigned level.}
    \label{fig:debias}
\end{figure}

In addition, we observe that recent progress in enhancing agentic capabilities of open-source models through reinforcement learning has been rapid, achieving high scores on the \gaia{} benchmark and substantially improving efficiency. This provides a highly promising direction for our future work.
\begin{table}[!h]
\centering
\setstretch{1.1}
\caption{Performance comparison of various base models on \gaia{} benchmark. All results are obtained using information retrieved from Google Search.}
\label{tab:crawler}
\vspace{-9pt}
\begin{tabular}{l|cccc}
\toprule
Model & Average & Level 1 & Level 2 & Level 3  \\
\midrule
WebShaper-32B & 53.3 & 69.2 & 50.0 & 16.6 \\
WebShaper-72B & 60.1 & 69.2 & 63.4 & 16.6 \\
\midrule
GPT-4.1 & 55.8 & 66.0 & 58.1 & 26.9 \\
o4-mini & 66.1 & 79.2 & 68.6 & 30.8 \\
Claude-3.7-sonnet & 68.3 & 83.1 & 70.6 & 30.8 \\
Claude-4-sonnet & 75.2 & 86.6 & 77.9 & 42.3 \\
\bottomrule
\end{tabular}
\end{table}

\begin{table}[!h]
\centering
\setstretch{1.2}
\caption{Performance of various search engine settings on \gaia{}. Note that `Multi-Source' refers to combine the above search engines.}
\vspace{-9pt}
\label{tab:search}
\begin{tabular}{l|cccc}
\toprule
Search Method & Average & Level 1 & Level 2 & Level 3  \\
\midrule
Google        & 75.2 & 86.6 & 77.9 & 42.3 \\
Bing          & 58.8 & 62.3 & 64.0 & 34.6 \\
DuckDuckGo    & 70.9 & 81.1 & 73.3 & 42.3 \\
Multi-Source  & 68.5 & 75.5 & 73.3 & 38.5 \\
\bottomrule
\end{tabular}
\end{table}

\noindent \textbf{Search Engine}
Web search is significant for LLM-agents to obtain external information beyond their knowledge boundaries. However, we found that different search engines and search engine service APIs have substantial impacts on task performance. We employed Google, Bing, DuckDuckGo, and their aggregated results (Multi-Source) respectively. As shown in Table~\ref{tab:search}, Agent with Google achieved the highest score of 75.2. Beyond differences in search result entries, a possible explanation is that Google (supported by SerpAPI) provides more fine-grained filtering condition settings, including date, location,  specific categories etc. For the GAIA benchmark, such fine-grained conditional filtering is essential.

\subsubsection{Level Debias}
The difficulty level assignments for tasks in the GAIA dataset rely as a proxy on the number of steps and tools used by our annotators when crafting the questions. However, there exists a gap between human behavior and machine behavior, where tasks that are simple for humans—such as visual recognition and browser operations—are indeed more challenging for machines. Therefore, we have reclassified them into four levels based on problem-solving success rates under our agent system, as shown in Fig~\ref{fig:debias}.

%% file: conclusion.tex
\section{Conclusion}
We present a unified framework for building effective generalist agents through the integration of heterogeneous agent paradigms, hierarchical memory, and a validated tool suite. Our design demonstrates that ensemble methods combining \texttt{Plan–Execute} and \texttt{ReAct} agents achieve both reliability and adaptability, while structured communication and layered memory maintain continuity across extended interactions. By curating essential tool categories—search, code execution, and multimodal parsing—we ensure factual grounding and reproducibility remain central to agent performance. On the \gaia{} benchmark, our approach achieves competitive results against both proprietary and open-source frameworks, establishing a new standard for robust, reproducible generalist agents.
We identify three promising directions for future agent research. First, \emph{dynamic self-improvement} through reinforcement learning and test-time scaling may enable ensembles to evolve coordination strategies beyond static voting mechanisms. Second, \emph{autonomous tool evolution} could allow agents to generate and refine their own tools, reducing manual engineering overhead~\cite{qiu2025alita}. Third, \emph{cross-domain transfer} through modular frameworks may enable planners to adapt seamlessly to new environments while preserving stable worker capabilities~\cite{owl2025}. These trajectories point toward agents that are not only benchmark-accurate but also resilient, adaptive, and truly general-purpose in real-world applications.

%% file: acknowledge.tex
\section{Contributions}
\begin{multicols}{1}

\textbf{JingDong}
\begin{itemize}
    \item Jiarun Liu
    \item Shiyue Xu
    \item Shangkun Liu
    \item Yang Li
    \item Wen Liu
    \item Min Liu
    \item Xiaoqing Zhou
    \item Hanmin Wang
    \item Shilin Jia
    \item zhen Wang
    \item Shaohua Tian
    \item Hanhao Li
    \item Junbo Zhang    
    \item Yongli Yu
    \item Peng Cao
\end{itemize}

\textbf{Tongji University}
\begin{itemize}
    \item Haofen Wang
\end{itemize}
\end{multicols}

%% file: appendix.tex
\newpage
\beginappendix

\section{Details of tools}
\label{app:tools}

\noindent \textbf{Search Tools} 
External search is essential for agent systems to extend knowledge boundaries, and we have implemented several fine grain search tools as follows:

\resizebox{0.85\textwidth}{!}{
\begin{promptbox}[Search Tools]{jingdongred}
    \textbf{Web Search}
    \begin{itemize}
    \item \textbf{Google Search}
    \item \textbf{Bing Search}
    \item \textbf{DuckDuckGo Search}
    \item \textbf{Integrated Search}
    \end{itemize}
    \textbf{Github Search}
    \begin{itemize}
        \item \textbf{Repository Search}
        \item \textbf{Issue Search}
        \item \textbf{PR Search}
        \item \textbf{Releases Search}
    \end{itemize}
    \textbf{Arxiv Search}
    \begin{itemize}
        \item \textbf{Advanced Search}
    \end{itemize}
    \textbf{Wiki Search}
    \begin{itemize}
        \item \textbf{Wikipedia Search}
    \end{itemize}      
\end{promptbox}
}

\noindent \textbf{Parsing Tools}
The correct parsing of files is a prerequisite for the Agent system to effectively utilize the information obtained. We have implemented a wealth of parsing tools as follows:

\resizebox{0.85\textwidth}{!}{
\begin{promptbox}[Parsing Tools]{jingdongred}
    \textbf{File Parsing}
    \begin{itemize}
        \item \textbf{PDF Tool} 
        \item \textbf{Doc Tool}
        \item \textbf{Text Tool}
        \item \textbf{Image Tool}
        \item \textbf{OCR Tool}
        \item \textbf{Audio Tool}
        \item \textbf{PDB Tool}
        \item \textbf{HTML Tool}
        \item \textbf{Zip Tool}
    \end{itemize}
    \textbf{Page Parsing}
    \begin{itemize}
        \item \textbf{Webpage Tool}
        \item \textbf{Archived Page Tool}
        \item \textbf{Wiki Page Tool}
        \item \textbf{Youtube Page Tool}
    \end{itemize}
\end{promptbox}
}

\noindent \textbf{Youtube Tools}
Without using the multimodal video mode, we have implemented multiple tools to capture different content of YouTube videos separately:

\newpage
\resizebox{0.85\textwidth}{!}{
\begin{promptbox}[Parsing Tools]{jingdongred}
    \textbf{Fetch Tool}
    \begin{itemize}
        \item \textbf{Video Introduction Tool}
        \item \textbf{Frame Screenshot Tool}
        \item \textbf{Subtitle Tool}
        \item \textbf{Audio Tool}
    \end{itemize}
\end{promptbox}
}

\noindent \textbf{Browswer Tools}
For some tasks that require interaction with web pages, we directly load the mcp tool provided by playwright.